\documentclass{article}

\usepackage{microtype}
\usepackage{graphicx}
\usepackage{subfigure}
\usepackage{booktabs} 

\usepackage{hyperref}

\usepackage[accepted]{icml2024}

\usepackage{amsmath}
\usepackage{amssymb}
\usepackage{mathtools}
\usepackage{amsthm}

\usepackage[capitalize,noabbrev]{cleveref}

\theoremstyle{plain}

\theoremstyle{definition}

\theoremstyle{remark}

\usepackage[textsize=tiny]{todonotes}

\icmltitlerunning{Using Images to Find Context-Independent Word Representations in Vector Space}

\begin{document}

\twocolumn[
\icmltitle{Using Images to Find Context-Independent Word Representations in Vector Space}




\begin{icmlauthorlist}
\icmlauthor{Harsh Kumar}{yyy}
\end{icmlauthorlist}

\icmlaffiliation{yyy}{Indian Institute of Science, Bengaluru, India}

\icmlcorrespondingauthor{Harsh Kumar}{harshkumar@iisc.ac.in}

\icmlkeywords{Machine Learning, ICML}

\vskip 0.3in
]



\printAffiliationsAndNotice{} 

\begin{abstract}
Many methods have been proposed to find vector representation for words, but most rely on capturing context from the text to find semantic relationships between these vectors. We propose a novel method of using dictionary meanings and image depictions to find word vectors independent of any context. We use auto-encoder on the word images to find meaningful representations and use them to calculate the word vectors. We finally evaluate our method on word similarity, concept categorization and outlier detection tasks. Our method performs comparably to context-based methods while taking much less training time.
\end{abstract}

\section{Introduction}
Using vector space representations for words is a widely studied area in natural language processing. Many methods have been proposed to finding semantically meaningful vector representations for words, including the bag-of-words model, skip-grams \cite{Mikolov2013EfficientEO}, character n-grams \cite{10.1162/tacl_a_00051}, matrix-factorization methods \cite{article}, \cite{article_rohde}, \cite{Bullinaria2007ExtractingSR}, \cite{DBLP:journals/corr/LebretL13} and RNN architectures \cite{luong-etal-2013-better}.

However, nearly all these methods treat word tokens as individual entities and attempt to form semantic similarity between their representations by capturing some context from the text. As a result, the meaning of any given word becomes limited to the diversity of contexts seen in the dataset. In other words, the learned embedding of a word is a static point in the vector space, and any semantic relationship between the word vectors is limited to the contexts seen in the dataset.

This work presents our study on representing words based on their dictionary definitions. We believe that (potentially multiple) definitions of a term can fully explain its meaning across contexts. This is consistent with traditional methods in pedagogy where students are often learn definitions of new terms along with their example usage.

Since there are a limited number of words in any vocabulary, other words are needed to explain the meaning of a given word. The ``definition" is thus a recursive relationship between words. To deal with this issue, we employ the pedagogical approach of using images to describe word meanings. For this, we create a custom image dataset for words encountered in the definitions of vocabulary words. We then use auto-encoders to find semantic representations for these images in a sequence to get the final word representation.

As we shall see, our method gives a comparable performance with context-based methods while being computationally cheaper, taking only ten hours for training.

We list our contributions as follows:
\begin{itemize}
    \item We advocate the usage of definitions for finding word vector representations instead of relying on diverse contexts to capture their meanings.
    \item We use images to represent the definition terms for the words and use auto-encoders to find hidden representations for these images. The hidden representations of these images are then appended together in a sequence to find the final representation of the original word.
    \item We release our custom dataset which consists of 5,00,000+ images for 1,15,000+ terms occurring in vocabulary and their definitions.
\end{itemize}

The rest of the paper has been organized as follows. In Section 2, we list previous works done in the field of word representations. In Section 3, we explain our approach including the custom dataset preparation and the architecture of auto-encoder used. In Section 4, we provide training details and the set of tasks on which we evaluate our method. In Section 5, we explain our observations and finally, in Section 6, we conclude our paper.

\section{Related Work}
The representation of words in the machine learning literature has a long history. Early methods represented words as one-hot vectors. This method did not capture any semantic relationship between words. A related method was the bag-of-words (BoW) representation. The bag-of-words approach created a vocabulary by tokenizing all the text, followed by assigning a vector of vocabulary length to each sentence. Each element in the vector corresponds to the frequency of words in the sentence. The TF-IDF improved upon the one-hot encoding by weighting the importance of words in documents based on their frequency in the text. However, it still treated words independently without capturing any semantic relationship.

\cite{10.5555/944919.944966} proposed neural-network architecture where a feedforward neural network with a linear projection layer was used to learn jointly the word vector representation and a statistical language model. \cite{Mikolov2009NeuralNB} proposed another neural network architecture where the word vectors learned using a hidden layer are used to train the language model.

\cite{Mikolov2013EfficientEO} take the neural-network-based representation a step further and propose model architectures for computing continuous vector representations (embeddings) that capture semantic relationships. They propose two models: CBOW and Skip-grams. CBOW predicts a word given its context and Skip-gram predicts the context given the word. \cite{pennington-etal-2014-glove} propose a global log-bilinear model that combines the matrix factorization and local context window methods. It creates word embeddings by leveraging word-word co-occurrence statistics from a text.

\cite{10.1162/tacl_a_00051} propose a representation of word vectors as a sum of character n-grams to preserve the morphology of words. Their approach works better on a dataset of rare words for word similarity and analogy tasks. \cite{Luong-etal:conll13:morpho} propose a novel RNN architecture capable of building representations for morphologically complex words from their morphemes. \cite{DBLP:journals/corr/Chen17aa} propose to represent a document as a simple average of word embeddings to capture the semantic meaning of documents during learning. \cite{DBLP:journals/corr/RoyGMJ16} use word vectors to deal with information retrieval. They represent documents and queries as sets of word-embedded vectors for indexing and scoring documents.
\cite{o-etal-2018-word} calculate word vector representations using a knowledge-based graph to generate the context of an ambiguous word. This improves the performance of the learned vector representations in word-sense disambiguation tasks.

\cite{maas-etal-2011-learning} argue that vector-based approaches to semantics can model rich lexical meanings but fail to capture sentiment information. Their proposal alleviates this problem. \cite{erk-pado-2008-structured} propose a structured vector space model to compute word vector representations to identify the meaning of word occurrences which can vary widely according to context. \cite{DBLP:journals/corr/VilnisM14} advocate for using density-based distributed embeddings for word representations instead of mapping words as a point in vector space. 

\cite{Klein_2015_CVPR} deal with the problem of associating a sentence with an image using Fisher vectors. \cite{7966071} use auto-encoders based on extreme learning to find word representations on a word-context matrix.

\section{Proposed Solution}
Our proposed solution is based on a simple approach employed in pedagogy. Visual examples are often observed to be effective when teaching a new term or concept to students and students the meaning of a concept is often related with some visual representation.

The meaning of a word is often explained using a sequence of other words (definitions). For clarity, we call the words appearing in the definition of a word as \textit{definition-terms} set. Each word in the \textit{definition-terms} also needs an explanation, which is another sequence. For a limited vocabulary, this inevitably leads to circular dependencies among the words.

\textbf{word} -- (meaning) ${\longrightarrow}$ \textbf{sequence of words}

To avoid this issue, we use images to represent the words found in the definitions. We call the list of these images as that word's \textit{image-set}. The collection of all image-sets for all the words in the vocabulary leads to a custom image dataset. We train an auto-encoder model on this image dataset to find meaningful representations. Representations for each image in the \textit{image-set} are appended to get the final representation of the word.

We explain the dataset preparation and the auto-encoder architecture in the following sections.

\subsection{Dataset Preparation}
We use the pre-trained BERT model's vocabulary as our base vocabulary. The meanings for these words are obtained using definitions for them available in dictionaries at Project Gutenberg. Any word found in the definitions, but not in the base vocabulary are also added to our final list of words. The final set of words forms our final vocabulary.

\begin{figure}
    \centering
    \includegraphics[width=0.5\linewidth]{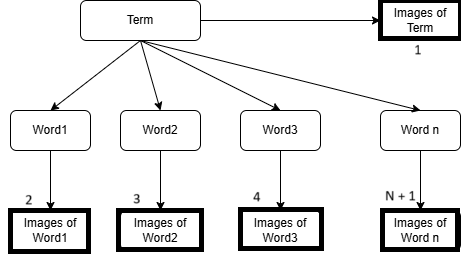}
    \caption{Creation of custom dataset. The images of the original word and the definition terms form the \textit{image-set} of the given word. The subscripts indicate the sequence.}
    \label{fig:overall_approach}
\end{figure}

Our dataset consists of images for terms in the final vocabulary in such a way that the \textit{image-set} of a given word consists of images for the words in its \textit{definition-terms} set. The images in the \textit{image-set} is organized in the same sequence as the words in the \textit{definition-terms}. The overall approach is shown in Figure \ref{fig:overall_approach} and is explained next.

\begin{itemize}
    \item For each term in the base vocabulary, we find its definition using a dictionary. We use the dictionaries available at Project Gutenberg and Wiktionary for the same. The new terms encountered in the definition are included in the final vocabulary. Our final vocabulary has 1,15,458 terms.
    \item We use the CommonCrawl dump image links and DuckDuckGo's image search API to search for openly available images for terms in the final vocabulary. The image set for each term consists of the images of the given term and images for the terms in the definition in the same sequence as they appear. \newline
    For example, if the definition of a given word \textbf{word} is given by the terms \textbf{w1} \textbf{w2} \textbf{w3} \textbf{w4} \textbf{w5}, then the image-set of \textbf{word} consists of list of images of \textbf{word}, \textbf{w1}, \textbf{w2}, \textbf{w3}, \textbf{w4} and \textbf{w5} in the exact sequence. We carefully select five images for each term to account for the fact that a given term can have multiple meanings.
    \item We limit the definition for each word to a maximum of 19 terms. We append empty tokens (represented by blank images) to the definition if their size is below the limit. As such, each word is represented by a set of $(19 + 1)*5 = 100$ images in a specific sequence. 
    
    Please note that common occurring words such as \textit{is}, \textit{are}, \textit{also} etc. are not considered. However, the emphatic words, question words, conjunctions and punctuations are not removed because these may change the tone and meaning of a sentence.
    
\end{itemize}

\subsection{Auto-Encoder}
\begin{figure}
    \centering
    \includegraphics[width=1.0\linewidth]{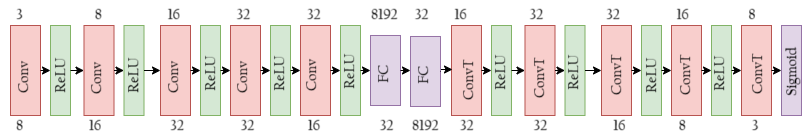}
    \caption{The Auto-encoder architecture used in our method to get the latent representation for images. The top subscripts show the number of input channels whereas the bottom subscripts show the number of output channels. Each conv and convT layer has 3x3 kernel with stride = 1 and padding = 1}
    \label{fig:autoencoder}
\end{figure}

We train an autoencoder model to find meaningful representations for the images in the dataset. The model consists of a five-layer encoder-decoder architecture with a latent state size of 32. Each encoder layer consists of a Convolution-ReLU block with convolution having a kernel size of 3x3, stride = 1, and padding = 1. Each decoder layer consists of a ConvTranspose2d-ReLU pair (the last layer has a sigmoid instead of the ReLU activation) with a kernel size of 3x3, stride = 1, and padding = 1 for the ConvTranspose2d layer. The overall architecture is shown in Figure \ref{fig:autoencoder}.

The input images are reshaped to 32x32 and the model finds a 32-dimensional vector representation. The vector representations of all the images in the word's image-set are appended to calculate the final 3200-dimensional vector representation for the given word.

\begin{figure}
    \centering
    \includegraphics[width=0.8\linewidth]{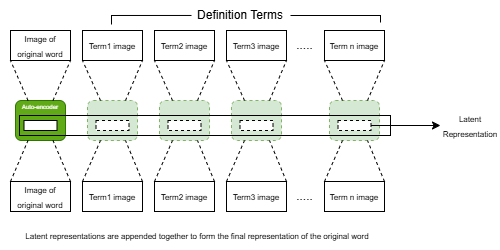}
    \caption{The auto-encoder is used to get the latent representation of images. These latent representations for definition terms are appended in a sequence to get the final representation of the original word}
    \label{fig:latent_representation_images}
\end{figure}

\section{Experiments}

\subsection{Training details}
We trained our auto-encoder model on a custom dataset of 5,77,290 images corresponding to 1,15,458 terms of our final vocabulary. Each image is resized to 32x32 pixels. We use binary-cross-entropy loss to measure the image reconstruction error. 

The autoencoder model is initialized with random weights and trained for 25 epochs. We use the Adam optimizer, with an initial learning rate of 0.00215. The learning rate is halved every 5 epochs. The training takes 10 hours on an NVIDIA Ampere GPU with 8GB GDDR6X.

\subsection{Evaluation Methods}
We conduct extensive evaluation experiments on various word semantic similarity, Concept Categorization, and Outlier word detection tasks. We evaluate our proposed method on several benchmarks whose details are listed as follows:
\subsubsection{Word Semantic Similarity}
The task of word semantic similarity is based on the idea that the distances between the words in the embedding space follow closely to the actual semantic distances evaluated using human judgments. We use the cosine distance between the word vectors to represent the semantic distance between their meanings. Our benchmarks include WordSim-353 \cite{Finkelstein:2002:PSC}, MC-30 \cite{doi:10.1080/01690969108406936}, RG-65 \cite{10.1145/365628.365657}, RW \cite{luong-etal-2013-better}, SimVerb-3500 \cite{DBLP:journals/corr/GerzVHRK16}, SimLex-999 \cite{DBLP:journals/corr/HillRK14}, MTurk-287 \cite{10.1145/1963405.1963455}, Verb-143 \cite{baker2014unsupervised}

\subsubsection{Outlier Word Detection}
The task is to identify a semantically anomalous word in an already-formed cluster. We evaluate our method on 8-8-8 \cite{camacho-collados-navigli-2016-find} and WordSim-500 \cite{DBLP:journals/corr/BlairMB16} datasets.

\subsubsection{Concept Categorization}
The task is to evaluate the quality of word representations clustered into subsets belonging to different categories. For example, the words \textit{chair}, \textit{bench}, \textit{bed} should be in a distinct cluster as compared to \textit{car}, \textit{bus} and \textit{truck}. We evaluate our method on the Battig and Montague \cite{Battig_Montague_1969}, Almuhareb and Poesio \cite{almuhareb-poesio-2004-attribute}, BLESS \cite{baroni-lenci-2011-blessed}, and ESSLLI-2008 \cite{esslli-2008s} benchmarks.

\section{Results}
We compare our proposed method with other word-embedding models in Table \ref{tab:table_word_similarity}. A similarity score is obtained using the cosine similarity. We compute spearman's rank correlation coefficient between the score and the human judgments. The data for other models were taken from \cite{DBLP:journals/corr/abs-1901-09785} since their study uses the Wiki2010 dataset for training (number of tokens in this dataset is comparable to number of images in our custom dataset). We observe that our approach gives comparable performance to other models while taking less training time.

Table \ref{tab:table_outlier_detection} lists down the accuracy for outlier word detection datasets. We observe comparable performance with other models.

We model the concept categorization as a clustering task and report v-measure scores for the benchmarks. The v-measure score is the harmonic mean between the homogeneity and completeness metrics and is independent of the labels assigned to the clusters. Table \ref{tab:table_concept_categorization} lists down v-measure scores for concept categorization datasets. We observe decent v-measure scores for all the datasets which indicates that the word representations help achieve a good clustering quality.

\section{Conclusion and Future Work}
In this work, we discussed a simple yet effective approach to finding semantically sound word vector representations. We employed an auto-encoder to find latent representations for images corresponding to words and their definition terms and appended those latent representations sequentially to obtain the find word vector representation.

We also observed finding such vector representations take a relatively small amount of time due to the smaller size of the auto-encoder model. At the same time, the obtained word vectors show decent performance on word similarity, concept categorization, and outlier detection tasks.

We want to point out that the main challenge with this approach is the creation of a custom dataset and picking relevant images for each term in the vocabulary. As such, one should expect the quality of the word vectors to be dependent on the selected images to represent the word. At the same time, it can be argued that this approach is independent of finding appropriate length context windows and using large text datasets which is pervasive with other models.

An interesting work would be to evaluate the effectiveness of our approach on machine translation tasks because a given object may be described using different words across different languages, however, images for those objects remain the same.

\onecolumn
\begin{table}[]
    \centering
    \begin{tabular}{ccccccccc}
        \hline
             & WS-353 & MC-30 & RG-65 & MTurk-287 & RW & SimLex-999 & Verb-143 & Simverb-3500 \\
        \hline
        \hline
        SGNS & 0.716 & 0.81 & 0.79 & 0.67 & 0.46 & 0.39 & 0.45 & 0.28 \\
        \hline
        CBOW & 0.64 & 0.74 & 0.81 & 0.67 & 0.43 & 0.37 & 0.40 & 0.24 \\
        \hline
        GloVe & 0.59 & 0.74 & 0.75 & 0.61 & 0.32 & 0.32 & 0.36 & 0.17 \\
        \hline
        FastText & 0.64 & 0.76 & 0.77 & 0.63 & 0.46 & 0.35 & 0.35 & 0.21 \\
        \hline
        ngram2vec & 0.74 & 0.85 & 0.79 & 0.66 & 0.45 & 0.42 & 0.47 & 0.32 \\
        \hline
        Dict2vec & 0.69 & 0.80 & 0.85 & 0.60 & 0.49 & 0.41 & 0.18 & 0.41 \\
        \hline
        Our Approach & 0.72 & 0.70 & 0.69 & 0.61 & 0.36 & 0.23 & 0.30 & 0.22 \\
        \hline
    \end{tabular}
    \caption{Spearman rank correlation on word similarity tasks. Data for all the other models are taken from \cite{DBLP:journals/corr/abs-1901-09785}}
    \label{tab:table_word_similarity}
\end{table}

\begin{table}[]
    \centering
    \begin{tabular}{ccc}
    \hline
            & 8-8-8 & WordSim-500  \\
        \hline
        \hline
         SGNS & 57.81 & 11.25 \\
         \hline
         CBOW & 56.25 & 14.02 \\
         \hline
         GloVe & 50.0 & 15.09 \\
         \hline
         FastText & 57.81 & 10.68 \\
         \hline
         ngram2vec & 59.38 & 10.64 \\
         \hline
         Dict2vec & 60.94 & 11.03 \\
         \hline
         Our Approach & 52.25 & 13.06 \\
         \hline
    \end{tabular}
    \caption{Accuracy of different word embedding models on outlier detection datasets. Data for all the other models are taken from \cite{DBLP:journals/corr/abs-1901-09785}}
    \label{tab:table_outlier_detection}
\end{table}

\begin{table}[]
    \centering
    \begin{tabular}{cc}
    \hline
        Dataset & V-Measure \\
        \hline
        \hline
        AP & 0.51 \\
        \hline
        Battig \& Montague & 0.76 \\
        \hline
        BLESS & 0.69 \\
        \hline
        ESSLI-2008 & 0.78 \\
        \hline
    \end{tabular}
    \caption{Performance of our approach on different concept-categorization datasets in terms of V-measure}
    \label{tab:table_concept_categorization}
\end{table}

\twocolumn[]

\bibliography{example_paper}
\bibliographystyle{icml2024}
\end{document}